\documentclass[sigconf]{acmart}
\usepackage{subcaption}
\AtBeginDocument{%
  }

\copyrightyear{2025}
\acmYear{2025}
\acmDOI{}
\acmConference[]{}{}{}

\renewcommand\footnotetextcopyrightpermission[1]{} 
\acmISBN{}

\def\@authornotemark{} 
\begin{document}

\title{Meme$\mathbb{CMD}$: An Automatically Generated $\mathbb{C}$hinese $\mathbb{M}$ulti-turn $\mathbb{D}$ialogue Dataset with Contextually Retrieved Memes}


\author{Yuheng Wang$^{\dagger}$, Xianhe Tang$^{\dagger}$, Pufeng Huang$^{\dagger}$}
\affiliation{%
  \institution{Wuhan University}
  \state{Wuhan}
  \country{China}
}
\thanks{$^{\dagger}$indicates equal contribution.}

\renewcommand{\shortauthors}{Yuheng Wang, Xianhe Tang, Pufeng Huang.}

\begin{abstract}
Memes are widely used in online social interactions, providing vivid, intuitive, and often humorous means to express intentions and emotions. Existing dialogue datasets are predominantly limited to either manually annotated or pure-text conversations, lacking the expressiveness and contextual nuance that multimodal interactions provide.To address these challenges, we introduce \textbf{MemeCMD}, an automatically generated \textbf{C}hinese \textbf{M}ulti-turn \textbf{D}ialogue dataset with contextually retrieved memes. Our dataset combines a large-scale, MLLM-annotated meme library with dialogues auto-generated by dual agents across diverse scenarios. We introduce a retrieval framework and adaptive threshold to ensure contextually relevant, naturally spaced meme usage. Experiments demonstrate the effectiveness of our approach in generating contextually appropriate and diverse meme-incorporated dialogues, offering a scalable and privacy-preserving resource for advancing multimodal conversational AI. Our code is available at \href{https://github.com/Nahtreom/MemeCMD}{https://github.com/Nahtreom/MemeCMD}.
\end{abstract}

\ccsdesc[500]{Computing methodologies~Discourse, dialogue and pragmatics}
\ccsdesc[500]{Computing methodologies~Natural language generation}
\ccsdesc[300]{Computing methodologies~Visual content-based indexing and retrieval}

\keywords{Multimodal Large Language Model, Dual Agent, Meme Retrieval, Chinese Dialogue, Automatic Generation}

\begin{teaserfigure}
  \begin{center}
  \includegraphics[width=0.7\textwidth]{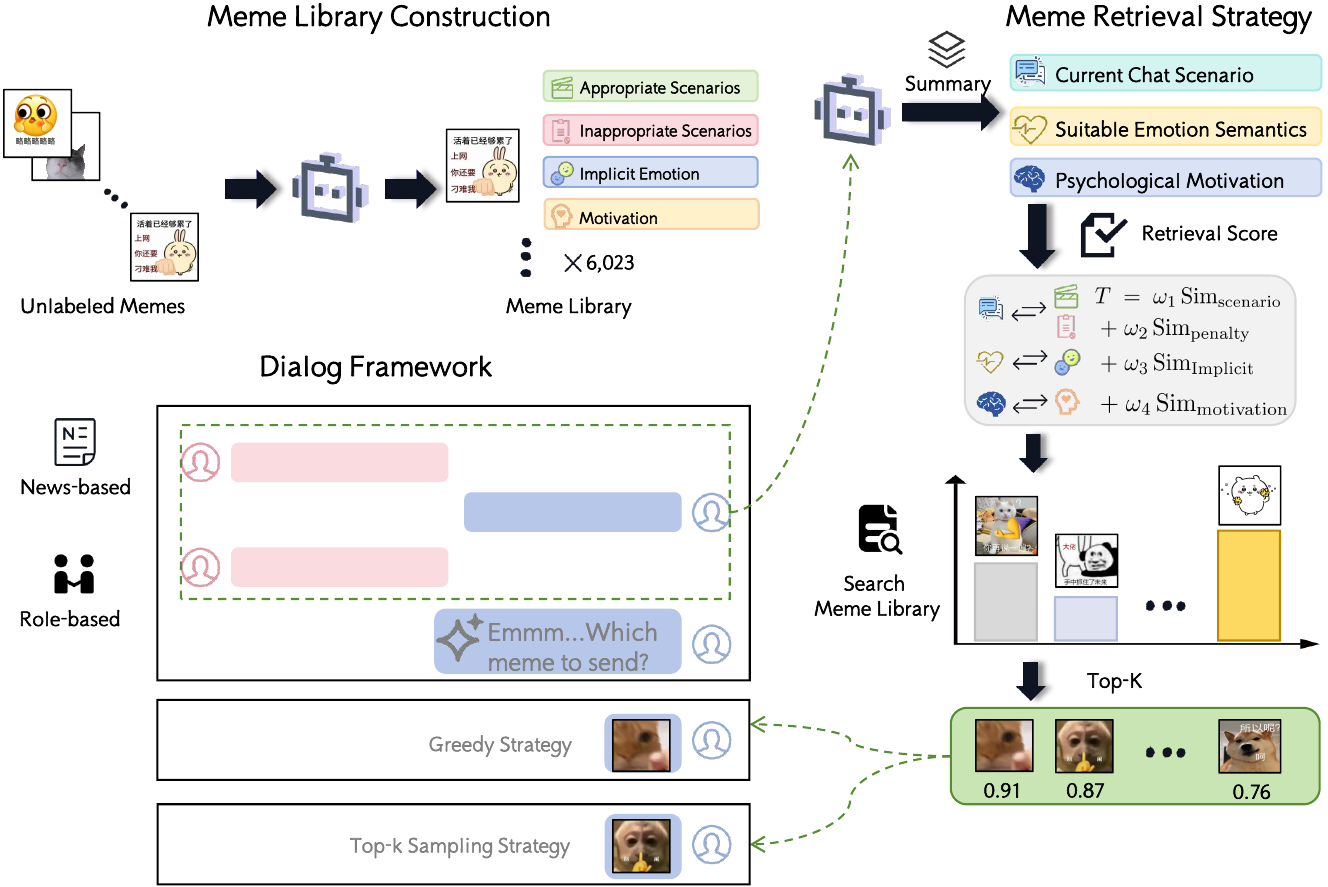}
  \caption{Overview of MemeCMD. The system consists of three major components: (1) Meme Library Construction, where unlabeled memes are annotated with scenario appropriateness, implicit emotion, and psychological motivation to build a labeled meme library; (2) Dialog Framework, which models either news-based or role-based conversations and selects memes via greedy or Top-$K$ sampling strategies; and (3) Meme Retrieval Strategy, which ranks memes by computing a retrieval score based on the similarity between the current chat scenario and meme annotations across three dimensions: scenario, emotion, and motivation.}
    \end{center}
  \label{fig:teaser}
\end{teaserfigure}


\maketitle

\section{Introduction}
As various online communication platforms continue to proliferate, online interactions have become an integral component of modern life\cite{zhang2024stickerconv}. A distinguishing feature of such interactions—compared to traditional communication—is the use of multimodal memes\cite{sharma2020semeval,Wang2025ANF}, which convey user intent in ways that go beyond plain text.


Owing to their unique visual expressiveness\cite{nilasari2018sticker}, memes can significantly enhance the engagement and entertainment value of a conversation\cite{DBLP:conf/naacl/BarbieriBRS18}, while also allowing speakers to convey their intentions in a concise and intuitive manner.


Existing research on meme retrieval has primarily focused on using memes as an enhancement to textual replies\cite{liu2022ser30k,zhao2023sticker820k}, or on retrieving memes based solely on the user’s current utterance\cite{fei2021towards}. However, this paradigm remains suboptimal in real-world scenarios. First, memes often convey redundant information\cite{pictory2024memes} when paired with text expressing the same intent—that is, users may choose to respond with a meme alone, rather than using it merely to reinforce a textual reply. Second, the intent behind a meme is frequently embedded within the broader conversational context\cite{zhao2025memereacon,roth2025context}. Relying solely on the current utterance to infer the intended meaning is often insufficient, as the meme's communicative function is tightly coupled with the surrounding dialogue.


Furthermore, our survey of existing meme-related resources and multi-turn dialogue datasets\cite{liu2024convbench} reveals two key gaps: the absence of well-annotated meme repositories\cite{zang2021photochat} and the scarcity of Chinese multi-turn dialogue datasets that include meme interactions. Additionally, existing dialogue datasets tend to have narrow topical scope and often face difficulties in mitigating privacy-related risks.


In response, we begin by manually filtering and integrating existing unannotated meme datasets. Building upon this, we employ MLLM to conduct large-scale automatic annotation, taking advantage of their ability to align visual and textual modalities. Each meme is annotated with information along four dimensions: (1) appropriate usage scenarios, (2) inappropriate scenarios, (3) implicit emotions and internet-specific meanings, and (4) the user’s psychological motivations and communicative intents. This process yields a comprehensive and well-annotated meme repository, termed the Meme Library, which contains a total of 6,023 memes.


Leveraging the Meme Library, we develop a pipeline framework with two main components: a dialogue generation module that uses either news events or character roles for initialization, and a meme retrieval module that selects memes based on the full conversational context and semantic relevance.


Concretely, we construct a dual-agent dialogue system, where conversations are initiated based on either news-based information or predefined role-based knowledge. After each turn, a Summary Agent extracts the underlying semantic representation of the dialogue context. This representation is then matched against the annotated dimensions in the Meme Library, and a custom retrieval function is used to score and rank candidate memes accordingly.


After computing the relevance scores, we extract the Top-$K$ meme candidates. To balance response quality and diversity, we propose two selection strategies: Greedy, which always selects the highest-scoring meme, and Sampling, which randomly selects one from the Top-$K$ candidates.

Since meme responses are not appropriate in every turn, we introduce a threshold mechanism: a meme is only sent if its score surpasses the predefined threshold. Additionally, to regulate the frequency of meme usage over multiple turns, we implement a Turn-Aware Meme Frequency Control via Adaptive Threshold Decay strategy, which dynamically lowers the threshold as the dialogue progresses.


The main contributions of this paper are summarized as follows:

\begin{itemize}
    \item We introduce Meme Library, a meme dataset comprising 6,023 images, each accompanied by high-quality and detailed annotations.
    \item We propose a pipeline framework for automatically generating multi-turn dialogues enriched with context-aware memes.
    \item We introduce the first automatically constructed Chinese multi-turn dialogue dataset featuring meme-enhanced responses, containing 34,758 dialogue turns in total.
\end{itemize}

\section{Related Work}
\subsection{Data Generation}

Traditional data acquisition is often constrained by high collection costs, quality fluctuations, and privacy risks. Generative data augmentation is increasingly becoming a solution. Researchers have made significant progress in multimodal fields, including image enhancement \cite{rombach2022high,wang2024multimodal,guo2025mambairv2,lan2024towards,luo20253denhancer,aydemir2024data}, text generation \cite{dai2025auggpt,coulombe2018text}, and audio synthesis \cite{zhou2024voxinstruct,tan2024naturalspeech}. Notably, NVIDIA's open-source Nemotron-4 340B model\cite{adler2024nemotron}, pre-trained using 98\% synthetic data , validates the effectiveness of generated data.

In our work, we utilize GPT-4\cite{achiam2023gpt} to generate data. This approach offers the advantages of broad scenario coverage and strong privacy protection. It not only supports diverse requirements concerning topics, identities, and styles but also avoids the risk of user information leakage associated with real data.

\subsection{Meme Retrieval Task}

With the evolution of expressive communication on online chat platforms, memes have become an important medium beyond text due to their effectiveness in conveying emotions and humor. Early representative work, such as SRS\cite{gao2020learning}, focused on recommending contextually appropriate memes to users based on multi-turn dialogue context. Subsequently, the MOD dataset \cite{fei2021towards} (Meme-incorporated Open-domain Dialogue) was constructed, comprising approximately 45k Chinese dialogues and 606k utterances.Recent studies include StickerInt \cite{liang2024reply}, which constructed a high-quality Chinese sticker dataset, and PerSRV \cite{chee2025persrv}, which explored the task of personalized meme recommendation.

In our work,we collect Chinese memes from the web and leverages GPT-4 to construct an automatically generated multi-turn dialogue dataset. Context-appropriate retrieval in diverse scenarios is achieved through joint modeling of meme annotations and conversational descriptions.

\section{Meme Library}


\textbf{Data Construction.} We began by surveying and collecting meme datasets on GitHub, ultimately selecting three highly starred Chinese meme repositories\cite{zhaooleeChineseBQB,getactivityEmojiPackage,llmredteamEmoVisualData}. Through manual curation, we then assembled an unlabeled collection of 6,023 images.

\noindent \textbf{MLLM Annotation.} Building on the curated unlabeled meme repository, we employed an automated annotation strategy using state-of-the-art MLLM to extract semantic insights\cite{zhu2023minigpt}. For each meme, the MLLM was prompted to generate four key pieces of information: (1) appropriate usage scenarios; (2) scenarios where the meme should not be used; (3) the implicit emotions and internet-specific meanings it conveys\cite{zhang2024visual,poria2018meld}; and (4) the sender’s underlying psychological motivations and communicative intentions.

Formally, given a meme image $ M_m \in \mathcal{M} $, where $ \mathcal{M} $ denotes the unlabeled meme repository, the tagging process can be defined as a mapping function:
\begin{equation}
f_{\text{MLLM}}: M_m \mapsto A_m = \{S_m^+, S_m^-, E_m, \Psi_m\}
\end{equation}

where $ f_{\text{MLLM}} $ represents the semantic tagging function implemented via the MLLM, and $ A_m $ is the structured annotation for image $ M_m $, consisting of:
\begin{itemize}
  \item $ S_m^+ $: appropriate usage scenarios,
  \item $ S_m^- $: inappropriate scenarios,
  \item $ E_m$: the implicit emotion and internet-specific meaning,
  \item $ \Psi_m $: the sender's psychological motivation and communicative intent.
\end{itemize}

Applying this function to all $ M_m \in \mathcal{M} $, we construct the annotated Meme Library $\mathcal{A}$:
\begin{equation}
\mathcal{A} = \{(M_m, A_m)\}_{m=1}^{N_m} = \{(M_m, f_{\text{MLLM}}(M_m))\}_{m=1}^{N_m}
\end{equation}

\begin{figure}
  \centering
  \begin{subfigure}[b]{0.45\linewidth}
    \includegraphics[width=\linewidth]{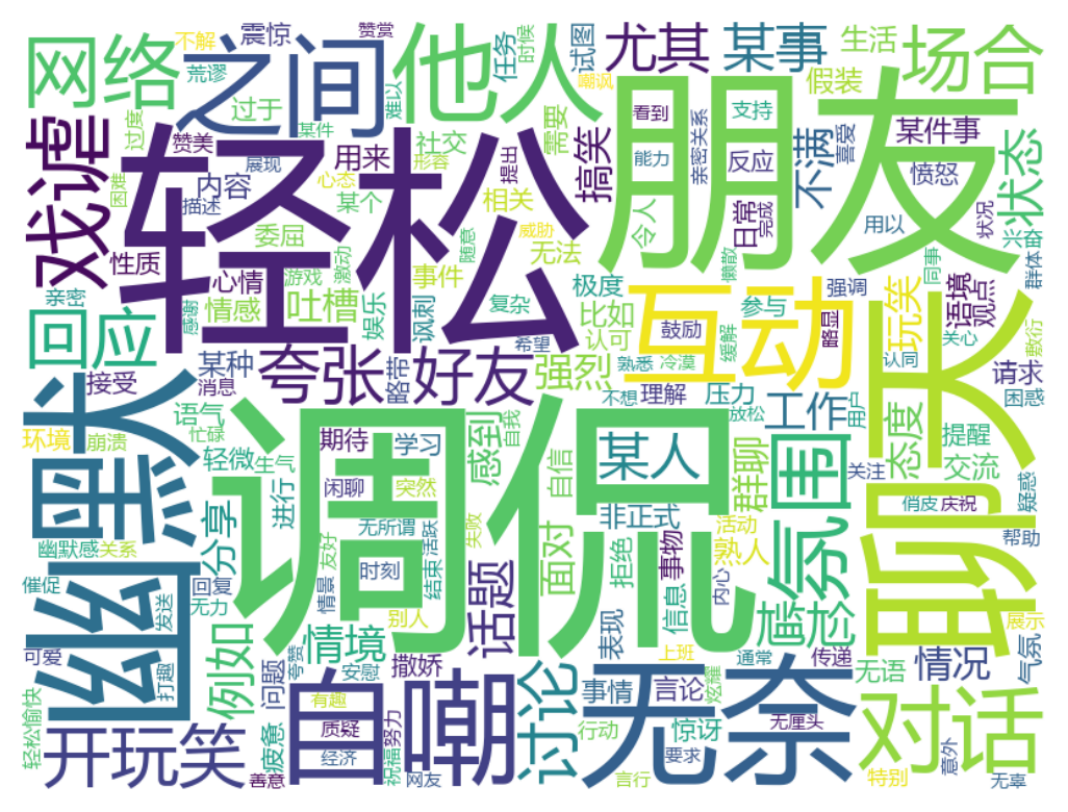}
    \caption{$S^+$: Suitable Scenarios}
  \end{subfigure}
  \begin{subfigure}[b]{0.45\linewidth}
    \includegraphics[width=\linewidth]{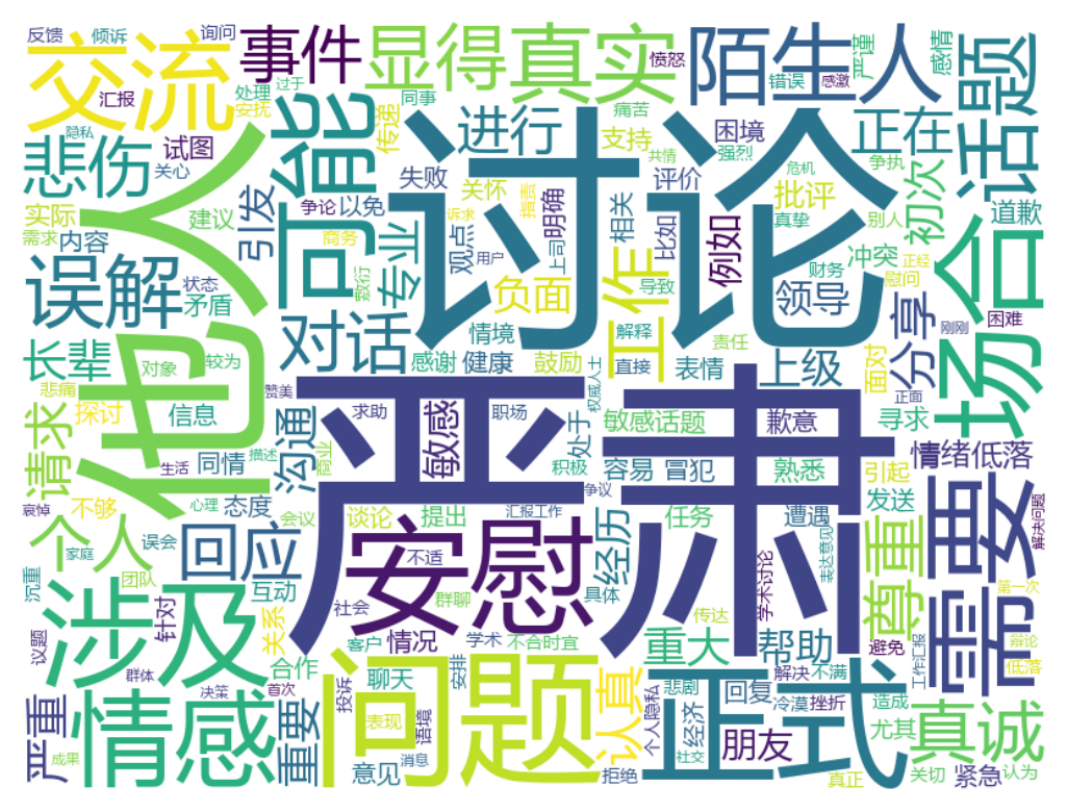}
    \caption{$S^-$: Forbidden Scenarios}
  \end{subfigure}

  \begin{subfigure}[b]{0.45\linewidth}
    \includegraphics[width=\linewidth]{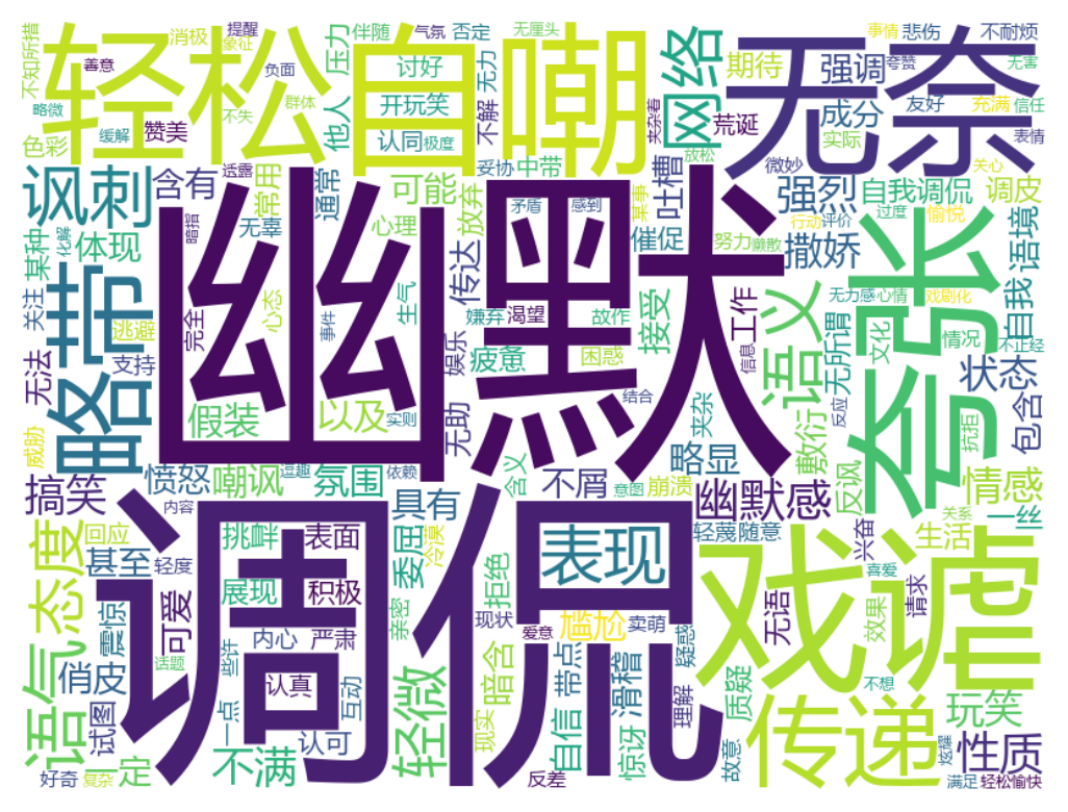}
    \caption{$E$: Emotions \& Semantics}
  \end{subfigure}
  \begin{subfigure}[b]{0.45\linewidth}
    \includegraphics[width=\linewidth]{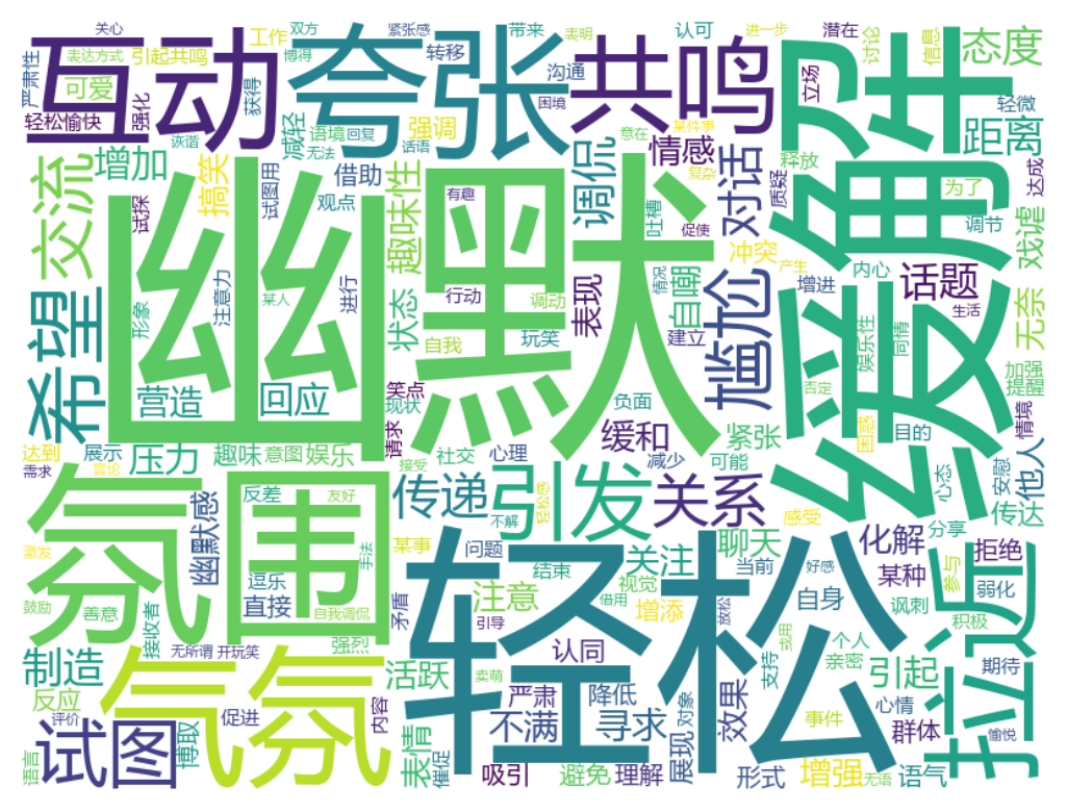}
    \caption{$\Psi$: User Motivation}
  \end{subfigure}

  \caption{Keyword Distributions Across Semantic Dimensions ($S^+, S^-, E, \Psi$) in the Meme Library Annotated by MLLM.}
  \label{fig:4pdfs}
\end{figure}

\noindent \textbf{Dataset Analysis.} 
We perform a comprehensive analysis of the constructed Meme Library. Specifically, we extract salient keywords from each annotated dimension $S^+, S^-, E, \Psi$. As shown in Figure~\ref{fig:4pdfs}, the word cloud visualizations reveal that the MLLM-generated annotations encompass a wide spectrum of topics, reflecting various facets of everyday life. Notably, the emotional dimension is not limited to humor and sarcasm, but also includes expressions such as avoidance and misunderstanding, highlighting the nuanced and diverse affective landscape captured by the memes. Taken together, these results demonstrate that the Meme Library constitutes a rich, semantically grounded, and structurally complete resource for downstream applications.

\section{Methodology}
\subsection{Dialog Framework}

Our framework supports two types of dialogue initialization (\emph{cold start}):  
(1) automatically extracted news information, and  
(2) manually defined role pairs, for which multiple prototypical scenarios are further generated by an agent.

Formally, for each dialogue session $i$ ($1 \leq i \leq N$), given the initial information $\mathcal{I}_i$ (either news-based or agent-generated scenario), the initialization process is defined as a mapping:
\begin{equation}
    f_\mathrm{init} : \mathcal{I}_i \mapsto \mathcal{C}_i = f_\mathrm{init}(\mathcal{I}_i)
\end{equation}
where $f_\mathrm{init}$ constructs the initial context $\mathcal{C}_i$ for session $i$.

The multi-turn dialogue for session $i$ is represented as:
\begin{equation}
    D_i = \{ u_{i,1}, u_{i,2}, \ldots, u_{i,T} \}
\end{equation}
where $u_{i,t}$ denotes the utterance at turn $t$ in session $i$.

At each turn $t$, the utterance is generated by the corresponding agent:
\begin{equation}
    u_{i,t} = A_{(t \bmod 2) + 1}\left(\mathcal{C}_i,\, u_{i,1:t-1}\right)
\end{equation}
where $A_{(t \bmod 2) + 1}$ denotes the agent for turn $t$, $\mathcal{C}_i$ is the initialization context for session $i$, and $u_{i,1:t-1}$ is the history for that session up to turn $t-1$.

The goal is to generate a coherent multi-turn dialogue for each session:
\begin{equation}
    D_i^* = \arg\max_{D_i} P(D_i \mid \mathcal{C}_i)
\end{equation}

The constructed dialogue dataset is then:
\[
\mathcal{D} = \left\{ (D_i, \mathcal{C}_i) \right\}_{i=1}^N
\]

\subsubsection{News-Based Cold Start}
Inspired by NaturalConv~\cite{wang2021naturalconv}, we implement a news-based cold start strategy by automatically collecting web news articles from diverse domains. Each news item is represented as a tuple $\mathcal{I}_i^\mathrm{news} = (\text{Topic}_i,\, \text{Title}_i,\, \text{Text}_i)$, where the topic is selected from one of six categories: Sports, Entertainment, Technology, Games, Education, and Health.

For each sampled news item $i$ ($1 \leq i \leq N_\mathrm{news}$), the initialization process is defined as a mapping:
\begin{equation}
    \mathcal{C}_i^\mathrm{news} = f_\mathrm{news}(\mathcal{I}_i^\mathrm{news})
\end{equation}
where $f_\mathrm{news}(\cdot)$ transforms the news tuple into a structured prompt for dialogue generation.

The agent $A_1$ then generates the first utterance $u_{i,1}$ for news item $i$:
\begin{equation}
    u_{i,1} = A_1(\mathcal{C}_i^\mathrm{news})
\end{equation}
Subsequent utterances $u_{i,t}$ ($t > 1$) are generated alternately by the dual agents, as described in Section~4.1.

Compared to traditional human-annotated dialogue datasets, this approach directly leverages news content as the initialization context, enabling fully automated and large-scale cold start dialogue generation over a diverse set of real-world topics.

\subsubsection{Role-Based Cold Start}

For role-based cold start, 23 representative role pairs are constructed by systematically combining four orthogonal relationship dimensions:
\begin{itemize}
    \item \textbf{Intimacy level:} stranger, acquaintance, close, very close
    \item \textbf{Dominance:} equal, A dominant, B dominant, mutual dependence
    \item \textbf{Age relation:} same age, A older, B older, generational difference
    \item \textbf{Primary scenario:} work, study, life, entertainment, special scenario
\end{itemize}

For each role pair $\mathcal{R}_i$ ($1 \leq i \leq N_\mathrm{role}$, e.g., supervisor-subordinate, classmates, landlord-tenant), an agent generates multiple low-similarity basic scenarios $\mathcal{S}_{i,j}$ ($1 \leq j \leq M_i$).

For each $(\mathcal{R}_i, \mathcal{S}_{i,j})$, the agent further generates semantically rich role information for both parties (background, personality, current state) and scenario details (relationship background, event context).

The aggregated initialization information for the $j$-th scenario of the $i$-th role pair is denoted as:
\[
\mathcal{I}_{i,j}^\mathrm{role} = (\text{role}_{i,A},\ \text{role}_{i,B},\ \text{scene}_{i,j})
\]

The initial dialogue context for each such scenario is constructed as:
\begin{equation}
    \mathcal{C}_{i,j}^\mathrm{role} = f_\mathrm{role}(\mathcal{I}_{i,j}^\mathrm{role})
\end{equation}
where $f_\mathrm{role}(\cdot)$ encodes the combined role and scene details into a structured prompt.

The agent $A_1$ then generates the first utterance $u_{i,j,1}$ for scenario $(i, j)$ based on $\mathcal{C}_{i,j}^\mathrm{role}$:
\begin{equation}
    u_{i,j,1} = A_1(\mathcal{C}_{i,j}^\mathrm{role})
\end{equation}
Subsequent utterances $u_{i,j,t}$ ($t > 1$) are generated alternately by agents as described previously.

The news-based approach is fully automated and highly extensible, but typically yields dialogue scenarios with limited diversity—mainly reflecting casual conversations between friends or acquaintances. In contrast, the role-based approach, while requiring some human involvement and being less scalable, substantially improves dialogue diversity and realism by encompassing a broad range of real-world social interactions.

\subsubsection{Dialog Agent Configuration}
The dialog agent framework employs dual agents, $\mathcal{A} = \{\text{A}, \text{B}\}$, who alternately generate utterances for each dialogue session $i$ ($1 \leq i \leq N$) over $T$ turns. The input prompt at each turn $t$ for session $i$, denoted $\mathcal{P}_{i,t}$, is dynamically constructed based on the stage of the conversation and the current context, as follows:

\textbf{Initial Turn ($t=1$):}  
The prompt is constructed with the complete initialization context for session $i$:
\begin{equation}
    \mathcal{P}_{i,1} = f_\mathrm{initPrompt}(\mathcal{C}_i,\varnothing, A_1)
\end{equation}

\textbf{Early Phase ($2 \leq t < 7$):}  
The prompt contains a concise summary of the context (e.g., news title or current character attributes), the accumulated dialogue history, and the current speaker:
\begin{equation}
    \mathcal{P}_{i,t} = f_\mathrm{earlyPrompt}(\mathcal{C}_{i,\mathrm{brief}},\, u_{i,1:t-1},\, A_t)
\end{equation}

\textbf{Middle Phase ($7 \leq t < 13$):}  
The prompt in this phase adopts special instructions designed to encourage more open-ended and topic-extending responses from the agent. It is formally constructed as:
\begin{equation}
    \mathcal{P}_{i,t} = f_\mathrm{middlePrompt}(\mathcal{C}_{i,\mathrm{brief}},\, u_{i,1:t-1},\, A_t)
\end{equation}

\textbf{Late Phase ($t \geq 13$):}  
All explicit initialization context is removed. The prompt consists solely of the current speaker identity and dialogue history:
\begin{equation}
    \mathcal{P}_{i,t} = f_\mathrm{latePrompt}(\varnothing,  u_{i,1:t-1},A_t)
\end{equation}
This phase encourages free-form and naturally evolving conversations.

At each turn, the agent response for session $i$ is generated as:
\begin{equation}
    u_{i,t} = A_t(\mathcal{P}_{i,t})
\end{equation}

All prompts enforce strict constraints on reply length , prohibit the use of emoji and repetitive sentence starters, and require colloquial, natural, and in-character responses.

This multi-stage, context-adaptive prompting strategy ensures that generated dialogues are initially well-grounded, then progressively become more open-ended, closely simulating realistic online conversational dynamics.

\subsubsection{Summary Agent}

Formally, for each dialogue session $i$ ($1 \leq i \leq N$) and each turn $t$ ($2 \leq t \leq T$), the summary agent generates a structured summary $S_{i,t}$, conditioned on the accumulated dialogue history $u_{i,1:t-1}$ and from the perspective of the next speaker $A_t$. The summary process can be defined as:
\begin{equation}
    f_\mathrm{summ} : (u_{i,1:t-1},\, A_t) \mapsto S_{i,t} = \left\{ \mathbf{S}_{i,t},\, \mathbf{E}_{i,t},\, \mathbf{\Psi}_{i,t} \right\}
\end{equation}
where $f_\mathrm{summ}$ is the summary function and $S_{i,t}$ is the structured output for session $i$ at turn $t$, consisting of:
\begin{itemize}
    \item $\mathbf{S}_{i,t}$: a concise description of the current chat scenario 
    \item $\mathbf{E}_{i,t}$: a concise summary of suitable emotional and internet-specific meanings for the next message 
    \item $\mathbf{\Psi}_{i,t}$: a concise summary of the sender's psychological motivation and communicative intent for sending a meme
\end{itemize}

Applying this summary agent to every session and turn, we obtain a set of structured dialogue summaries:
\begin{equation}
    \mathcal{S} = \left\{ \left( u_{i,1:t-1},\, A_t,\, S_{i,t} \right) \right\}_{i=1,t=2}^{N,T}
\end{equation}

\subsection{Retrieval Strategy}
First, we leverage MiniCPM-Embedding \cite{openbmb2024minicpm}to generate vector representations for the four annotated attributes of each meme in the Meme Library, as well as the three high-level labels produced by the Summary Agent. These representations are then projected into a unified semantic space to enable effective similarity computation and alignment. Next, we explore how to efficiently retrieve the most contextually appropriate memes from the large-scale Meme Library at any point in an ongoing multi-turn conversation.

\subsubsection{Meme Aligner}Based on the annotation information and the summarized multi-turn dialogue context, the Meme Aligner function is meticulously designed to integrate four key components: scenario matching, scenario penalization, implicit semantic matching., and motivation alignment. 

\noindent \textbf{Scenario Matching.} This module aims to measure the semantic similarity between the representation of the current dialogue scenario and the candidate meme’s scenario label. Scenario matching serves as a fundamental filtering step to ensure contextual appropriateness. It helps the system eliminate memes that may be emotionally relevant but contextually misplaced, which could otherwise lead to reduced communicative effectiveness or even misunderstanding. By aligning the dialogue context with the meme’s intended usage scenario, we ensure that the selected memes fit naturally within the ongoing conversation. Formally, given the current dialogue scenario vector $\mathbf{S}_{i,t}$ and a candidate meme $m$ with scenario vector $S_m^+$, the similarity score is computed as:

\begin{equation}
\alpha_{i,t,m} = \text{Sim}_{\text{scenario}}(\mathbf{S}_{i,t}, S_m^+) = \frac{\mathbf{S}_{i,t} \cdot S_m^+}{\|\mathbf{S}_{i,t}| \|S_m^+\|}
\end{equation}

\noindent \textbf{Scenario Penalization.} Similarly, we incorporate a scenario penalization strategy, which is equally critical—since, in many cases, it is preferable to withhold a meme rather than risk sending one that is contextually inappropriate. Specifically, for each meme candidate $m$, we compute its similarity to $S_m^{-}$, which denotes the negative scenario vector associated with meme $m$. The penalization score is defined as:

\begin{equation}
\delta_{i,t,m} = \text{Sim}_{\text{penalty}}(\mathbf{S}_{i,t}, S_m^{-}) = -\frac{\mathbf{S}_{i,t} \cdot S_m^{-}}{\|\mathbf{S}_{i,t}\| \|S_m^{-}\|}
\end{equation}

\noindent \textbf{Implicit Semantic Matching.} Memes often carry subtle expressions and multiple layers of meaning. Therefore, the Meme Aligner needs to leverage the inferred speaker intent at the current dialogue turn from the perspective of implicit semantics to assist meme selection. By uncovering and matching the underlying communicative purposes, we can prioritize memes that not only align with the surface content of the dialogue but also resonate at a deeper semantic and emotional level. This implicit semantic matching mechanism helps enhance the emotional resonance and communicative effectiveness of the recommended memes.Similarly, the implicit semantic matching score is defined as:
\begin{equation}
\beta_{i,t,m} = \text{Sim}_{\text{Implicit}}(\mathbf{E}_{i,t}, E_m) = -\frac{\mathbf{E}_{i,t}\cdot E_m}{\|\mathbf{E}_{i,t}\| \|E_m\|}   
\end{equation}

\noindent \textbf{Motivation Alignment.} In multi-turn dialogues, user utterances often reflect not only surface-level semantics but also deeper psychological motivations—such as seeking comfort, expressing empathy, or relieving stress. A key goal of the Meme Aligner is to leverage the inferred psychological needs of the speaker and align them with the underlying motivational signals conveyed by candidate memes. By incorporating this layer of motivational alignment, the system can more precisely identify memes that are emotionally effective and contextually appropriate\cite{10830638}, thereby enhancing both the emotional impact and communicative quality of the recommendations.
The motivational alignment score is computed as:
\begin{equation}
\gamma_{i,t,m} = \text{Sim}_{\text{motivation}}(\mathbf{\Psi}_{i,t}, \Psi_m) = \frac{\mathbf{\Psi}_{i,t} \cdot \Psi_m}{\|\mathbf{\Psi}_u\| \|\Psi_m\|}
\end{equation}

To achieve fine‑grained control over the meme matching task, Meme Aligner simultaneously considers the four core factors during retrieval and integrates them into a single optimization objective via a weighted sum. Specifically, the overall retrieval score $T_{i,t,m}$ for a query–meme pair can be expressed as: 

\begin{equation}
\label{eq:retrieval}
    T_{i,t,m} \;=\; 
\omega_1\, \alpha_{i,t,m}
\;+\;
\omega_2\, \delta_{i,t,m}
\;+\;
\omega_3\, \beta_{i,t,m}
\;+\;
\omega_4\, \gamma_{i,t,m}
\end{equation}


\begin{figure}[t]
    \centering
    \includegraphics[width=1\linewidth]{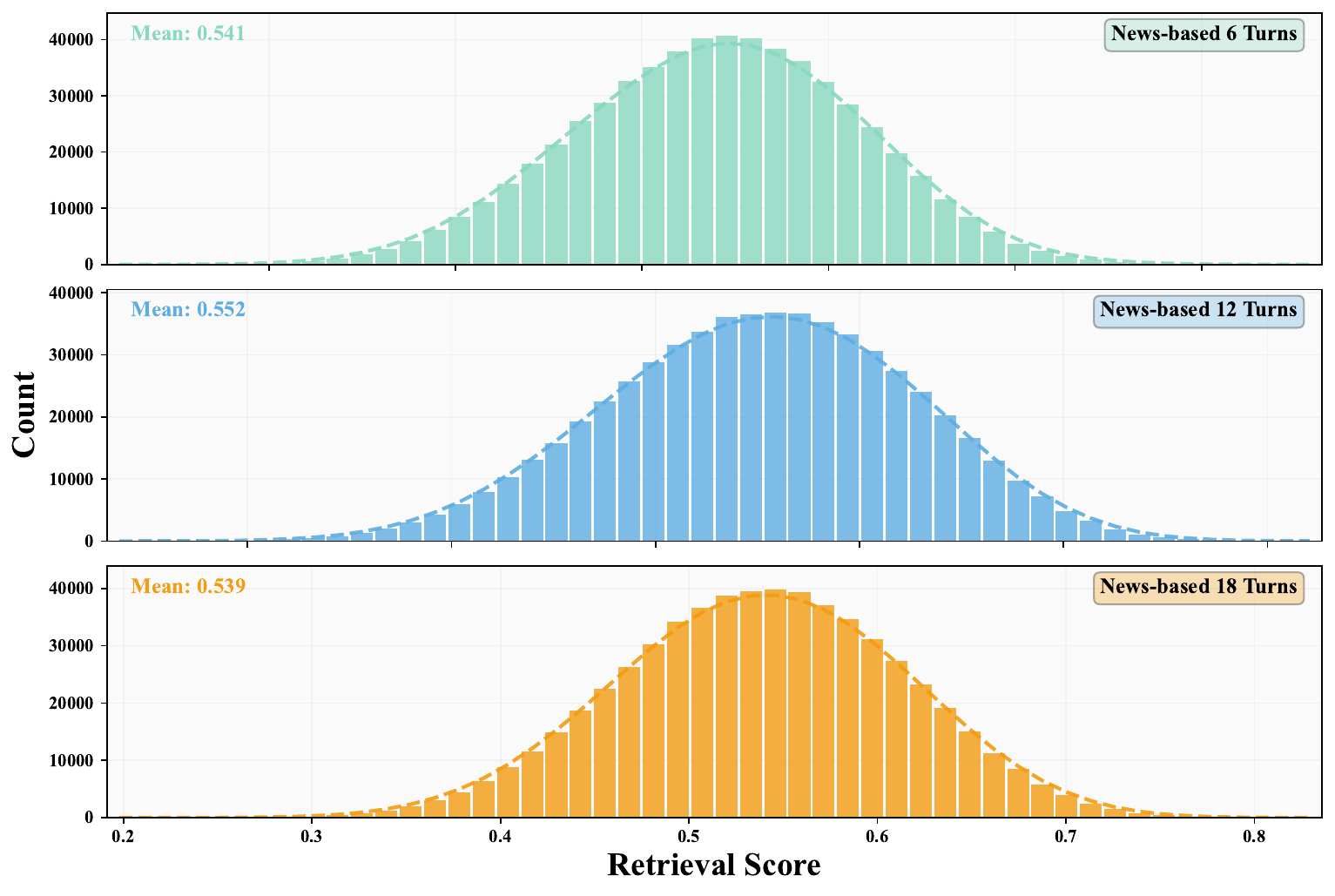}
    \caption{The distribution of Retrieval Scores in the Meme Library for the News-based dialogue dataset.}
    \label{Gate}
\end{figure}

After defining the overall retrieval score, we apply it to the dialogue dataset for meme retrieval. As shown in Figure~\ref{Gate}, the resulting score distribution closely resembles a normal distribution, which is highly consistent with our expectations. This suggests that the scoring function provides a smooth and discriminative signal across candidate memes, effectively separating highly relevant matches from less suitable ones. Overall, the result further confirms the effectiveness and robustness of our retrieval strategy in integrating the four semantic dimensions.

\begin{table*}[!tbp]
    \centering
    \caption{Experimental Results on News-Based and Role-Based Chinese Multi-Turn Dialogue Datasets. Bold formatting indicates the best metric for each dataset under the corresponding evaluation method, while underlining denotes the second-best metric. Across all datasets and methods, the random control group consistently yielded the lowest metrics.}
    \label{tab:experiment_result}
    \begin{tabular}{lccc|ccc}
        \hline
           \multirow{2}{*}{\textbf{Dataset}} & \multicolumn{3}{c}{\textbf{ LLM-as-a-Judge}} & \multicolumn{3}{c}{\textbf{Semantic Consistency}}\\
        \cline{2-7}
         & \textbf{Greedy} & \textbf{Sampling} & \textbf{Random} & \textbf{Greedy} &\textbf{Sampling} & \textbf{Random} \\
        \hline
       \rowcolor{gray!25}
       \textbf{News-based 6 Turns}    & \underline{86.66} \textcolor{teal}{\scalebox{0.7}{$\uparrow 3.23$}} & \textbf{87.10} \textcolor{teal}{\scalebox{0.7}{$\uparrow 3.67$}} & 83.43 & \textbf{68.49} \textcolor{teal}{\scalebox{0.7}{$\uparrow 0.10$}} & \underline{68.48} \textcolor{teal}{\scalebox{0.7}{$\uparrow 0.09$}} & 68.39 \\
        \textbf{News-based 12 Turns}   & \underline{89.45} \textcolor{teal}{\scalebox{0.7}{$\uparrow 1.65$}} & \textbf{89.94} \textcolor{teal}{\scalebox{0.7}{$\uparrow 2.14$}} & 87.80 & \textbf{68.71} \textcolor{teal}{\scalebox{0.7}{$\uparrow 0.07$}} & \underline{68.70} \textcolor{teal}{\scalebox{0.7}{$\uparrow 0.06$}} & 68.64 \\
                \rowcolor{gray!25}
        \textbf{News-based 18 Turns}   & \underline{90.33} \textcolor{teal}{\scalebox{0.7}{$\uparrow 0.12$}} & \textbf{90.68} \textcolor{teal}{\scalebox{0.7}{$\uparrow 0.47$}} & 90.21 & \textbf{68.74} \textcolor{teal}{\scalebox{0.7}{$\uparrow 0.10$}} & \underline{68.72} \textcolor{teal}{\scalebox{0.7}{$\uparrow 0.08$}} & 68.64 \\    
       \textbf{Role-based 6 Turns}    & \underline{88.82} \textcolor{teal}{\scalebox{0.7}{$\uparrow 4.65$}} & \textbf{88.93} \textcolor{teal}{\scalebox{0.7}{$\uparrow 4.76$}} & 84.17 & \textbf{68.10} \textcolor{teal}{\scalebox{0.7}{$\uparrow 0.21$}} & \underline{68.06} \textcolor{teal}{\scalebox{0.7}{$\uparrow 0.19$}} & 67.89 \\
        \rowcolor{gray!25}
        \textbf{Role-based 12 Turns }  & \textbf{91.35} \textcolor{teal}{\scalebox{0.7}{$\uparrow 2.67$}} & \underline{91.28} \textcolor{teal}{\scalebox{0.7}{$\uparrow 2.60$}} & 88.68 & \textbf{68.03} \textcolor{teal}{\scalebox{0.7}{$\uparrow 0.16$}} & \underline{68.01} \textcolor{teal}{\scalebox{0.7}{$\uparrow 0.14$}} & 67.85 \\
       \textbf{Role-based 18 Turns}   & \underline{91.42} \textcolor{teal}{\scalebox{0.7}{$\uparrow 0.58$}} & \textbf{91.56} \textcolor{teal}{\scalebox{0.7}{$\uparrow 0.72$}} & 90.84 & \textbf{68.03} \textcolor{teal}{\scalebox{0.7}{$\uparrow 0.15$}} & \underline{68.02} \textcolor{teal}{\scalebox{0.7}{$\uparrow 0.14$}} & 67.88 \\
        \hline
    \end{tabular}
\end{table*}

\begin{figure*}[!tbp]
    \centering
    \includegraphics[width=1\linewidth]{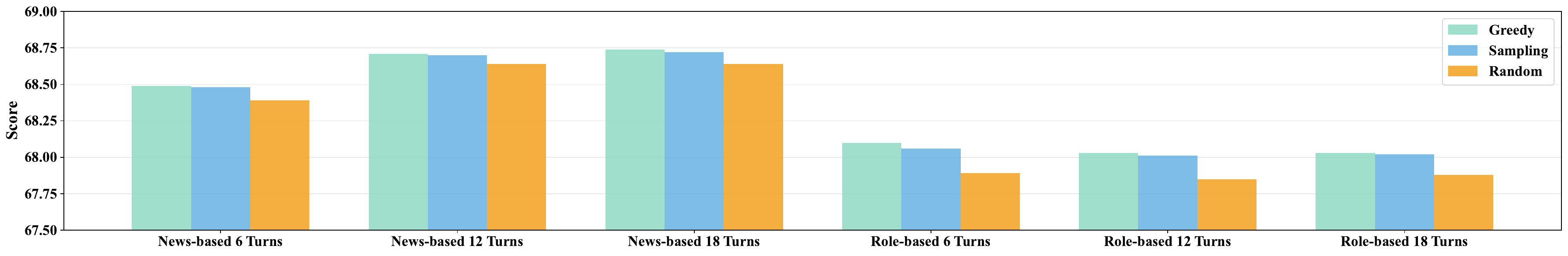}
    \caption{Cross-Modal Embedding-Based Semantic Consistency Evaluation Results demonstrate that both Greedy selection and Diversity-aware Sampling yield significantly higher semantic consistency scores than the random control group.}
    \label{fig:experiment_result}
\end{figure*}
\subsubsection{Turn-Aware Meme Frequency Control via Adaptive Threshold Decay}

In real-world online conversations, users rarely send memes or emojis in consecutive turns; instead, there is typically a natural pause between such events, preventing overuse and maintaining the expressiveness of memes. To emulate this behavior in our automatically generated dialogue, we introduce a turn-aware adaptive threshold decay mechanism for meme insertion.

After analyzing the retrieval score distribution, we observe that applying a fixed threshold can easily result in unrealistic meme frequency—either too sparse or too dense. Therefore, for each turn $t$ in dialogue session $i$, we adaptively adjust the meme-sending threshold based on the distance from the most recent prior meme event.

Let $\theta_0$ denote the base threshold, $\Delta$ the penalty term, and $\lambda$ the decay rate.  
Define $k_{i,t}$ as the number of turns since the last turn $t'$ ($t' < t$) in session $i$ where a meme was sent; if no meme has been sent before turn $t$, set $k_{i,t} = \infty$.

The adaptive threshold for meme sending at turn $t$ is then defined as:
\begin{equation}
\label{eq:turn_aware_decay}
    \theta_{i,t} = \theta_0 + \Delta\, e^{-\lambda k_{i,t}}
\end{equation}
where, by convention, $e^{-\lambda k_{i,t}} = 0$ when $k_{i,t} = \infty$, so that $\theta_{i,t} = \theta_0$ if no meme has ever been sent.

At each turn $t$, the retrieval score $T_{i,t,m}$ for the top candidate meme is compared to the current threshold $\theta_{i,t}$:
\begin{itemize}
    \item If $T_{i,t,m} > \theta_{i,t}$, a meme $M_{m}$ is sent at turn $t$;
    \item Otherwise, no meme is inserted for that turn.
\end{itemize}

This strategy penalizes meme deployment immediately after a recent meme was sent, with the penalty decaying exponentially as the gap increases. It ensures that meme occurrences are spaced more naturally throughout the conversation, closely mimicking human conversational behavior and improving the realism of our generated dialogue data.


\subsubsection{Meme Selector}


Next, Meme Selector directly rank all memes in the Meme Library based on their retrieval scores and select the Top‑$K$ candidates with the highest scores.

\begin{equation}
\{M_{(1)}, M_{(2)}, \dots, M_{(K)}\} = \operatorname{TopK}_{M \in \mathcal{M}} \big( T(M) \big)
\end{equation}


\noindent After obtaining the Top-$K$ candidate memes $\{M_{(1)}, M_{(2)}, \dots, M_{(K)}\}$, the Meme Selector offers two strategies to determine the final meme to be sent: Greedy selection: directly choose the top-1 meme with the highest retrieval score; Diversity-aware Sampling: randomly sample one meme from the Top-$K$ set to encourage output diversity, while maintaining overall semantic relevance.

\begin{equation}
M^* =
\begin{cases}
M_{(1)} & \text{(Greedy)} \\
\mathrm{Random}(\{M_{(1)}, \dots, M_{(K)}\}) & \text{(Sampling)}
\end{cases}
\label{eq:selector_strategy}
\end{equation}

\section{Experiments}
This section proposes several quantitative metrics to evaluate the effectiveness of different meme selection strategies. Given that meme usage must align with real conversational contexts, the ideal evaluation approach would be expert-driven assessments. However, this method is costly, poorly scalable, and highly subjective, leading to inconsistent evaluation results~\cite{gu2024survey}. Balancing reliability and scalability, we employ two complementary evaluation methods: the LLM-as-a-Judge paradigm and cross-modal embedding-based semantic consistency scoring.
\subsection{LLM-as-a-Judge Evaluation}
Recent studies indicate that LLMs optimized through supervised instruction fine-tuning and reinforcement learning from human feedback (RLHF), such as GPT-4, achieve over 80\% consistency with human preferences on subjective evaluation datasets. Thus, the LLM-as-a-Judge paradigm efficiently simulates human preferences while preserving interpretability~\cite{zheng2023judging}.

The multi-turn dialogues incorporating memes are presented to the MLLM as chat screenshots. The model evaluates each meme on a 0–100 scale across the following five dimensions:
\begin{itemize}
  \item \textbf{Semantic Consistency}: Alignment between the meme’s meaning and the conversational context.
  \item \textbf{Emotional Consistency}: Match between the meme’s sentiment and the contextual emotion.
  \item \textbf{Contextual Appropriateness}: Naturalness of meme usage within the current context, without awkwardness.
  \item \textbf{Humor Effect}: Whether the meme generates unexpected humor or sarcasm.
  \item \textbf{Conversational Coherence}: Whether the meme disrupts dialogue flow (e.g., interrupting topics or causing disconnection).
\end{itemize}

\subsection{Cross-Modal Embedding-Based Semantic Consistency Evaluation}
To explore scalable automated evaluation, we introduce a consistency metric based on image-text embeddings. This method quantifies the semantic relevance between a meme and the subsequent reply, serving as an indicator of contextual alignment~\cite{radford2021learning}. Since meme selection strategies rely solely on preceding dialogue, their semantic correlation with subsequent utterances reflects contextual adaptability.

We employ a MLLM to separately extract embeddings from meme images and their corresponding textual responses. The cosine similarity between these embeddings is then calculated and normalized onto a 0 \textasciitilde 100 scale:
\begin{equation}
  \label{eq:exper1}
\begin{aligned}
\text{CosSim}(\mathbf{A}, \mathbf{B}) = {}&\frac{\mathbf{A} \cdot \mathbf{B}}{\mathbf{||A||}\ \cdot  \mathbf{||B||}\
}\\
  \text{Score}_{[0,100]} ={}& \left( \frac{\text{CosSim} + 1}{2} \right) \times 100\\
  \end{aligned}
\end{equation}

However, current  MLLMs exhibit significant limitations in understanding emotional or implicit intent between memes and context. They often perform surface-level image analysis, struggling to capture communicative functions and affective nuances of memes, thereby limiting their efficacy in evaluating emotional consistency~\cite{zhao2025memereacon}.

\subsection{Experimental Settings}
In the LLM-as-a-Judge Evaluation, we employ the \textbf{chatgpt-4o-latest} API to score chat dialogues.
In the Cross-Modal Embedding-Based Semantic Consistency Evaluation, we utilize the \textbf{chinese-clip-vit-base-patch16} model for embedding extraction.

We evaluate three meme selection strategies:
\begin{itemize}
  \item \textbf{Greedy selection:} directly choose the top-1 meme with the highest retrieval score.
  \item \textbf{Diversity-aware Sampling:} randomly sample one meme from the Top-$3$ set to encourage output diversity, while maintaining overall semantic relevance.
    \item \textbf{Random:} Inserts a meme into each dialogue with 50\% probability, selected randomly from the Meme Library.
\end{itemize}
Among them, the threshold selection follows Equation~\ref{eq:turn_aware_decay}, with $\theta_0$ set to 0.7, $\Delta$ set to 0.2, and $\lambda$ set to 1.The Random group serves as a control to demonstrate the reliability of our evaluation strategy.

\begin{figure}[!tbp]
  \centering
  \begin{subfigure}[b]{0.32\linewidth}
      \includegraphics[width=\linewidth]{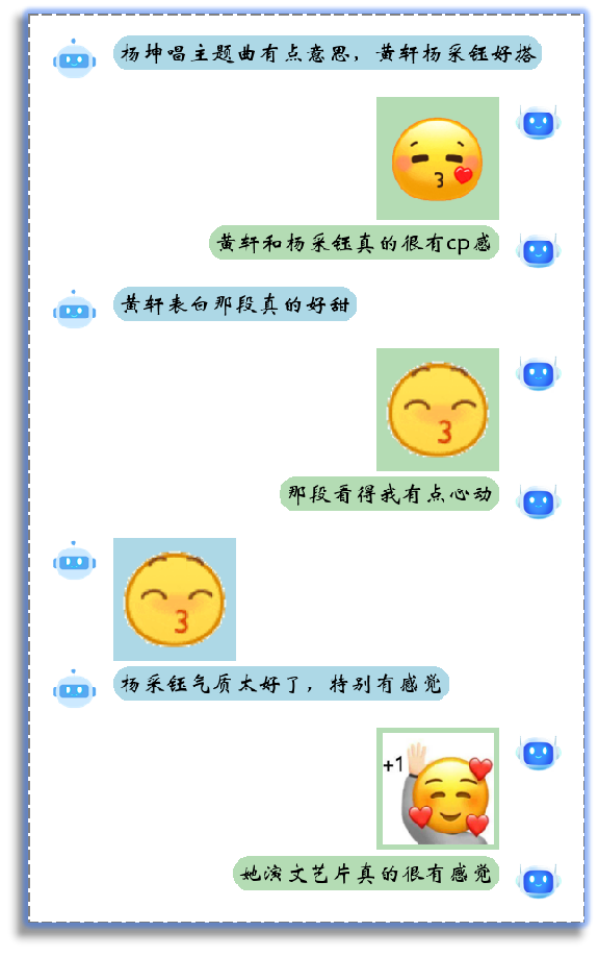}
    \caption{}
  \label{fig:4756a}
  \end{subfigure}
  \begin{subfigure}[b]{0.32\linewidth}
    \includegraphics[width=\linewidth]{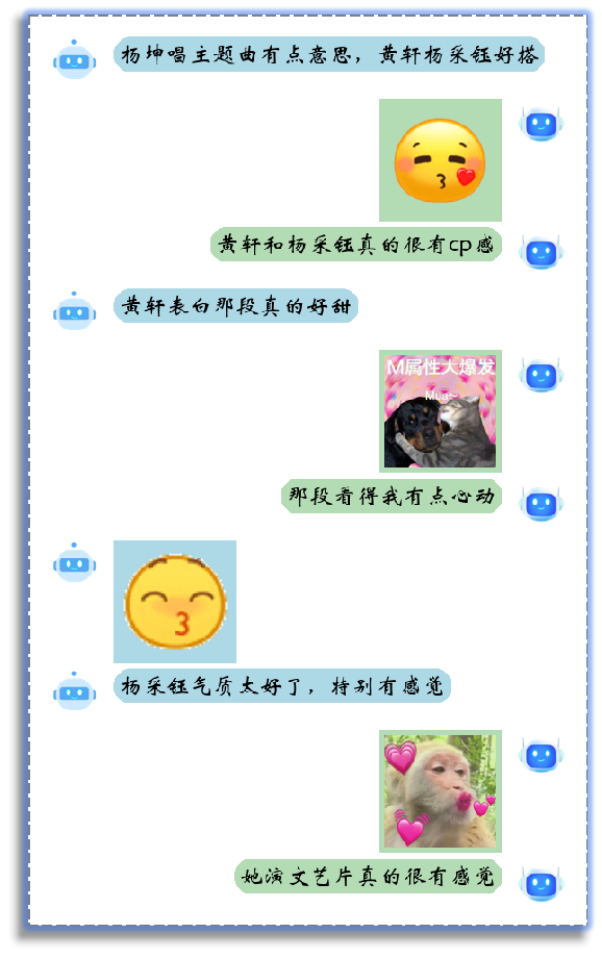}
    \caption{}
  \label{fig:4756b}
  \end{subfigure}
  \begin{subfigure}[b]{0.32\linewidth}
    \includegraphics[width=\linewidth]{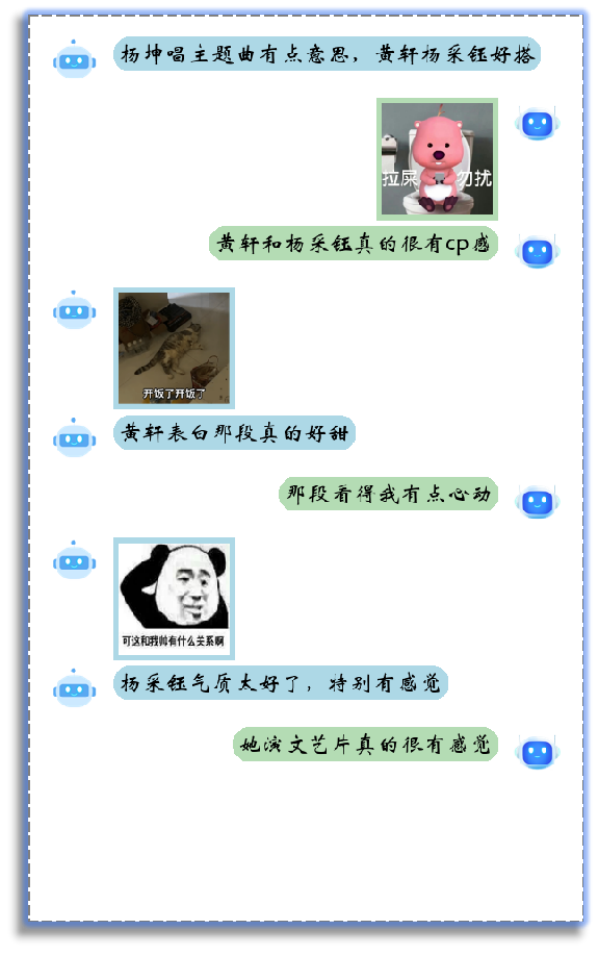}
    \caption{}
  \label{fig:4756c}
  \end{subfigure}

    \begin{subfigure}[b]{0.32\linewidth}
    \includegraphics[width=\linewidth]{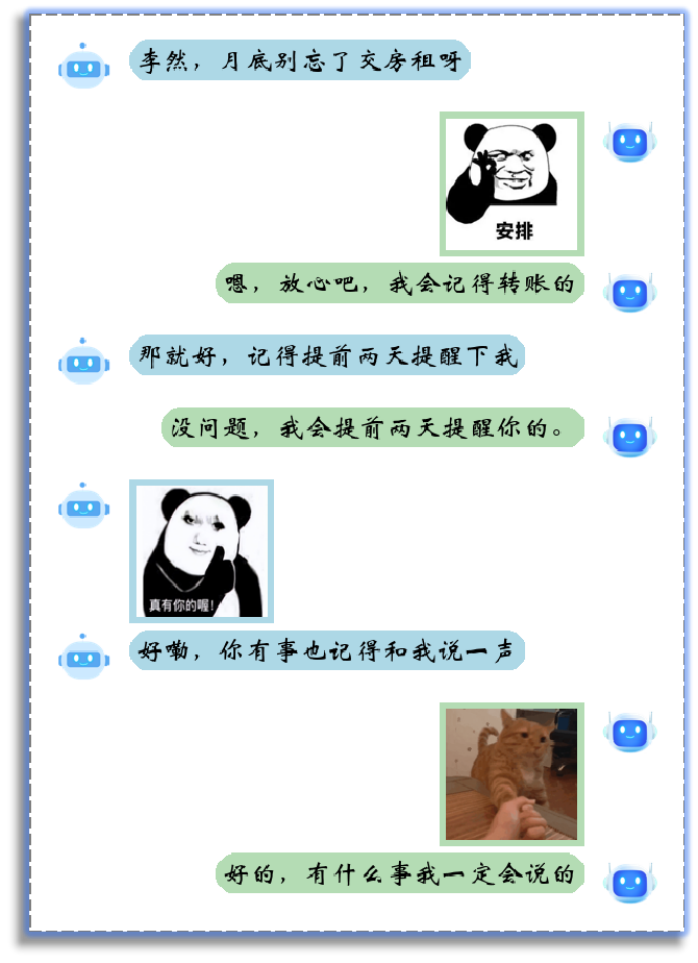}
    \caption{}
  \label{fig:0011a}
  \end{subfigure}
  \begin{subfigure}[b]{0.32\linewidth}
    \includegraphics[width=\linewidth]{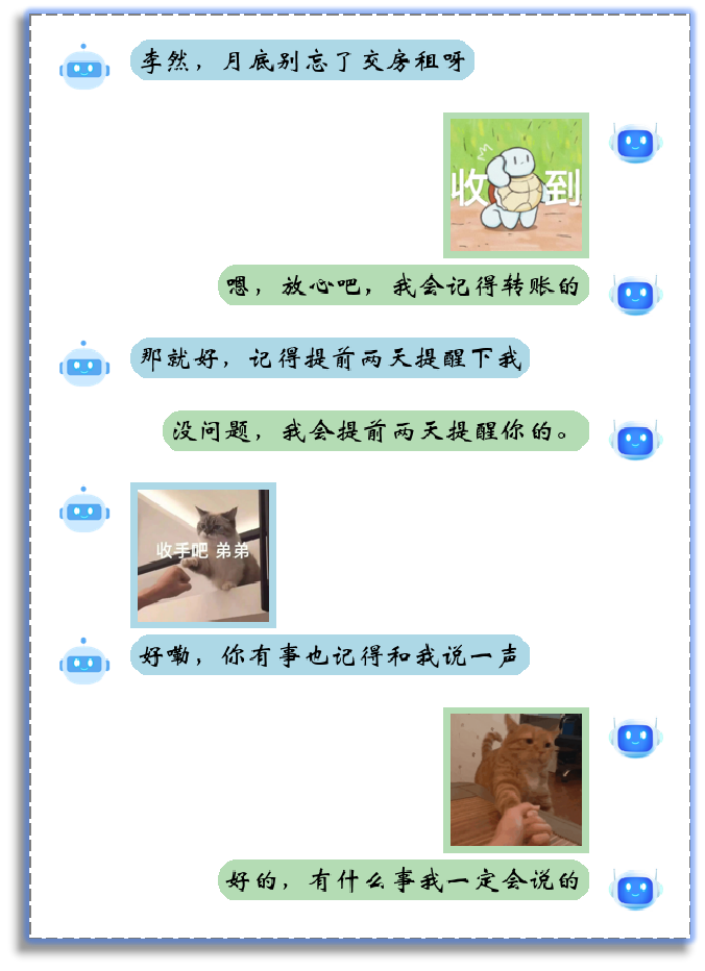}
    \caption{}
    \label{fig:0011b}
  \end{subfigure}
  \begin{subfigure}[b]{0.32\linewidth}
    \includegraphics[width=\linewidth]{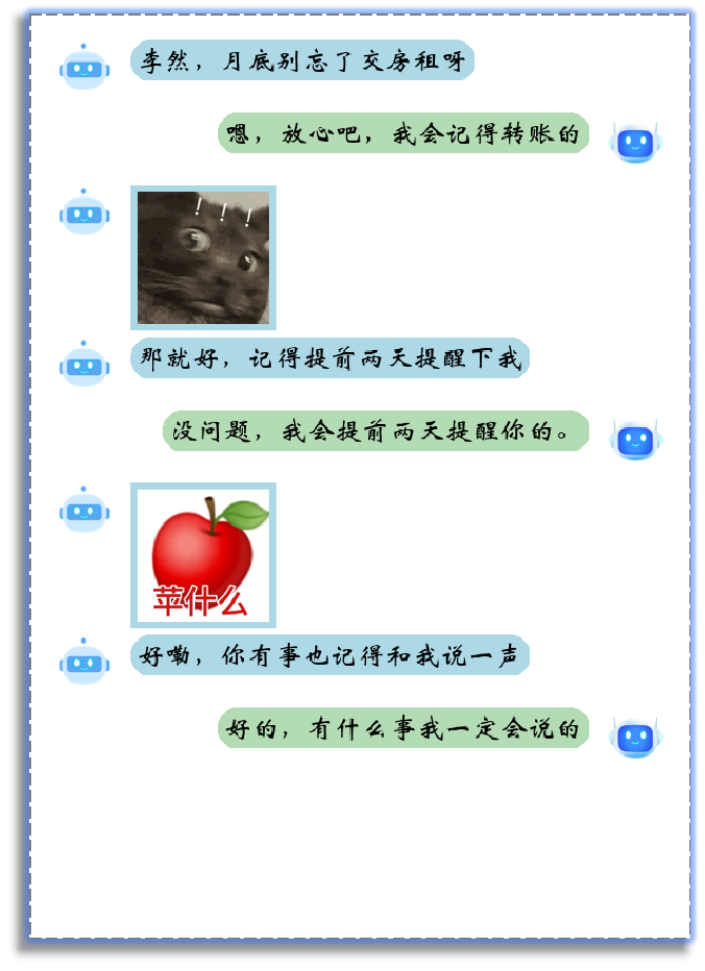}
    \caption{}
    \label{fig:0011c}
  \end{subfigure}
  \caption{Comparative Effectiveness of Multi-Strategy Meme Retrieval: (\subref{fig:4756a} \& \subref{fig:0011a}) \textbf{Greedy selection} strategy; (\subref{fig:4756b} \& \subref{fig:0011b}) \textbf{Diversity-aware Sampling} strategy; (\subref{fig:4756c} \& \subref{fig:0011c}) \textbf{Random Selection} baseline.}
  \label{fig:4756&0011}
\end{figure}

\subsection{Results Analysis}

\begin{table}[!tbp]
    \centering
    \rowcolors{2}{white}{gray!25}         
    \caption{Comparison of LLM-as-a-Judge and Cross-Modal Embedding-Based Semantic Consistency Scoring in Case Studies: The random control group demonstrates significantly lower scores than both Greedy selection and Diversity-aware Sampling methods.}
    \label{tab:case_study}
    \begin{tabular}{lcc}
        \hline
        {} & \textbf{LLM‑as‑a‑Judge} & \textbf{Semantic Consistency} \\
        \hline
        \textbf{Dialogue \ref{fig:4756a}} & \textbf{92.5} & \textbf{68.4} \\
        \textbf{Dialogue \ref{fig:4756b}} & \underline{91.5}         & \underline{68.3}         \\
        \textbf{Dialogue \ref{fig:4756c}} & 62.5         & 66.6         \\
        \hline
        \textbf{Dialogue \ref{fig:0011a}} & \underline{91.5}         & \underline{69.2}         \\
        \textbf{Dialogue \ref{fig:0011b}} & \textbf{92.5}& \textbf{69.8}\\
        \textbf{Dialogue \ref{fig:0011c}} & 62.5         & 68.3         \\
        \hline
    \end{tabular}
\end{table}

Experimental results show that retrieval-based strategies consistently outperform the random baseline across all evaluation dimensions and datasets. In practice, while the Greedy selection strategy selects the most contextually relevant meme, it often results in repetitive usage of the same image across turns. In contrast, the Diversity-aware Sampling strategy enhances meme diversity while maintaining contextual appropriateness. (as detailed in Table~\ref{tab:experiment_result} and Figure~\ref{fig:experiment_result}).



\noindent \textbf{Case Study.} Evaluation results for sample dialogues (Figure \ref{fig:4756&0011}) with corresponding scores (Table \ref{tab:case_study}) demonstrate that memes selected through our Greedy Selection and Diversity-Aware Sampling strategies consistently outperform those from the random selection baseline. This performance gap substantiates the empirical effectiveness of our proposed evaluation framework.

Further human evaluation reveals the contextual precision of our retrieval strategy, demonstrating effective affective alignment between multimodal expression and conversational context:
\begin{itemize}
\item In the dialogue shown in Figures \ref{fig:4756a}\textasciitilde\ref{fig:4756c}, the retrieved meme visually reinforces participants' appreciation and affection toward both the film and actors' performances
\item In the dialogue shown in Figures \ref{fig:0011a}\textasciitilde \ref{fig:0011c}, the retrieved meme accurately conveys acknowledgment intent ("received") while the "kitten handshake" image fosters a cordial atmosphere.
\end{itemize}

\section{Conclusion}


We introduce a method for automatically generating a Chinese Multi-turn Dialogue Dataset covering diverse conversational scenarios using GPT-4. This method simultaneously selects contextually appropriate memes for each dialogue turn. Our approach effectively addresses the challenges of inconsistent real-world data quality and privacy concerns inherent in dialogue datasets.

For meme selection, we propose a retrieval strategy that achieves high-quality, unsupervised meme matching. Additionally, we developed two novel evaluation methods to calculate retrieval scores in the absence of ground-truth annotations. By establishing random meme selection as a control group baseline, experimental results demonstrate the robustness and validity of our evaluation framework.

Future work will validate our retrieval strategy across diverse real-world dialogue datasets. We aim to leverage this approach as a scalable solution for expanding meme-retrieval dataset volumes,  enhancing multimodal dataset quality, and enriching LLM-generated responses with contextual diversity and engaging humor.

\bibliographystyle{ACM-Reference-Format}
\bibliography{sample-base}


\end{document}